\documentclass[sigplan,screen]{acmart}  %
\usepackage{multirow}

\usepackage{colortbl}
\usepackage{xcolor}

\AtBeginDocument{%
  }

\copyrightyear{2025}
\acmYear{2025}
\setcopyright{cc}
\setcctype{by}
\acmConference[EuroMLSys '25]{The 5th Workshop on Machine Learning and Systems }{March 30-April 3, 2025}{Rotterdam, Netherlands}

\acmBooktitle{The 5th Workshop on Machine Learning and Systems (EuroMLSys '25), March 30-April 3, 2025, Rotterdam, Netherlands}
\acmDOI{10.1145/3721146.3721949}
\acmISBN{979-8-4007-1538-9/2025/03}

\begin{document}

\title{$\beta$-GNN: A Robust Ensemble Approach Against Graph Structure Perturbation}

\author{Haci Ismail Aslan}
\email{aslan@tu-berlin.de}
\orcid{0000-0002-3647-5054}
\affiliation{%
  \institution{Technische Universität Berlin}
  \city{Berlin}
  \country{Germany}
}

\author{Philipp Wiesner}
\email{wiesner@tu-berlin.de}
\orcid{0000-0001-5352-7525}
\affiliation{%
  \institution{Technische Universität Berlin}
  \city{Berlin}
  \country{Germany}
}

\author{Ping Xiong}
\email{p.xiong@tu-berlin.de}
\orcid{0009-0009-1008-2138}
\affiliation{%
  \institution{Technische Universität Berlin}
  \city{Berlin}
  \country{Germany}
}

\author{Odej Kao}
\email{odej.kao@tu-berlin.de}
\orcid{0000-0001-6454-6799}
\affiliation{%
  \institution{Technische Universität Berlin}
  \city{Berlin}
  \country{Germany}
}

\renewcommand{\shortauthors}{Aslan et al.}

\begin{abstract}

Graph Neural Networks (GNNs) are playing an increasingly important role in the efficient operation and security of computing systems, with applications in workload scheduling, anomaly detection, and resource management.  However, their vulnerability to network perturbations poses a significant challenge.  We propose $\beta$-GNN, a model enhancing GNN robustness without sacrificing clean data performance. $\beta$-GNN uses a weighted ensemble, combining any GNN with a multi-layer perceptron.  A learned dynamic weight, $\beta$, modulates the GNN's contribution.  This $\beta$ not only weights GNN influence but also indicates data perturbation levels, enabling proactive mitigation. Experimental results on diverse datasets show $\beta$-GNN's superior adversarial accuracy and attack severity quantification.  Crucially, $\beta$-GNN avoids perturbation assumptions, preserving clean data structure and performance.

\end{abstract}



\maketitle

\section{Introduction}

GNN applications are rapidly expanding due to advancements in GNN generalization. 
By leveraging both node attributes and the topology, GNNs perform downstream tasks (i.e., node classification) more effectively than other models~\cite{rnnvsgnn}. 
Computing systems increasingly rely on reliable and robust GNNs to improve efficiency in operation as well as security, for example in network intrusion detection~\cite{intrusion}, workload scheduling~\cite{scheduling}, social networks~\cite{socialnetworks}, transportation systems~\cite{transportation}, or chip placement~\cite{nature_chip_design}. 
A key challenge to achieving this robustness lies in the demonstrated vulnerability of GNNs to adversarial attacks. Recent studies have substantiated these vulnerabilities, showing how attacks can manipulate the graph structure by adding adversarial noise to node features or manipulating edges via rewiring~\cite{survey, nettack, meta}.

This vulnerability has driven research into GNN robustness, with approaches ranging from model-centric enhancements to data-centric adversarial training. While these strategies aim to mitigate the impact of attacks, they often suffer from drawbacks, notably scalability issues~\cite{scalability}, which persist despite the advancements achieved in training GNNs at scale~\cite{gnnlab}. Furthermore, many existing methods focus on reacting to perturbations rather than detecting them. This reactive approach can lead to unnecessary graph cleaning operations, potentially distorting clean graphs and yielding suboptimal results. 

In this work, we introduce $\beta$-GNN, a solution that addresses these limitations. $\beta$-GNN enhances the robustness of GNNs by integrating any GNN model with a multi-layer perceptron (MLP) in a weighted ensemble. Unlike existing methods, $\beta$-GNN learns a dynamic weighting factor, denoted as $\beta$, which adjusts the contribution of the GNN model in the final prediction layer. This not only improves the model’s ability to withstand adversarial attacks but also provides an interpretable metric for quantifying the severity of perturbations in the data. Practitioners can leverage this information to take preventive actions when the data is highly perturbed, thus offering a proactive approach to GNN robustness. Our key contributions are as follows.

\begin{itemize}
    \item We propose $\mathbf{\beta}$-GNN, a flexible and modular framework applicable to any GNN architecture. We introduce $\mathbf{\beta}$ as a learned parameter that balances model performance and robustness, serving as a diagnostic tool for assessing graph perturbation severity.
    \item We conduct extensive experiments on both homophilic and heterophilic datasets, demonstrating that $\beta$-GNN achieves state-of-the-art results in terms of node classification accuracy under adversarial conditions.
    \item We provide a thorough analysis of the learned $\beta$ values, illustrating how they can be used to track the severity of attacks and guide response strategies. The reported results can be reproduced using our open-source implementation\footnote{\url{https://github.com/AslantheAslan/beta-GNN}}.
\end{itemize}

The remainder of this paper is structured as follows: In Section 2, we review the related work and motivation of this study. Section 3 introduces the $\beta$-GNN model architecture and its training process. Section 4 presents experimental results and analysis. Section 5 concludes the paper.

\section{Related Work}
Adversaries can subtly perturb graph structures or node features, leading to significant performance degradation in tasks like node classification and link prediction. In particular, poisoning attacks, where the graph is manipulated during the training phase, resulting in poor performance during inference, have been a primary focus of recent research. For instance, Nettack~\cite{nettack} introduced a targeted poisoning attack designed to modify both the graph structure and node features. The method seeks to alter a minimal number of edges and features to misclassify a specific node, without significantly disrupting the overall structure of the graph. Nettack formulates the attack as a bi-level optimization problem where the GNN is first trained on the poisoned graph, and the adversary then evaluates how modifications affect classification performance. An extended version of Nettack, called Metattack~\cite{meta}, on the other hand, proposes a more general and scalable poisoning attack by formulating the problem as meta-learning~\cite{meta-learning}, where the adversarial perturbations are optimized in a meta-gradient approach. Unlike Nettack, Metattack is untargeted and attacks the entire graph.

In response to vulnerabilities exposed by adversarial attacks, several defense strategies have been proposed to enhance the robustness of GNNs. These methods either aim to detect and remove the adversarial perturbations or to strengthen the GNN model itself. To filter out potential adversarial edges before training, GCN-Jaccard~\cite{gcnjaccard} calculates the Jaccard similarity between feature sets of connected nodes and removes edges with low similarity. This approach assumes that adversarial edges are more likely to connect dissimilar nodes, however, it causes problems if the adversarial influence is the opposite, or the data itself is heterophilic. Pro-GNN~\cite{prognn} adopts a two-step approach that combines graph structure learning with adversarial robustness. It formulates the defense as a joint optimization problem, denoising the graph using low-rank approximation and sparsity constraints while training a robust GNN. However, Pro-GNN may remove essential structural information along with perturbations, potentially degrading performance on the original graph. Similarly, since adversarial attacks mainly affect high-rank properties, constructing a low-rank graph via truncated singular value decomposition (TSVD)~\cite{TSVD} improves GNN robustness. This idea is further refined through reduced-rank topology learning, which preserves only the dominant singular components of the adversarial adjacency matrix to maintain the graph spectrum.

Alongside methods that mitigate adversarial impact on graphs, training-based approaches have also advanced GNN robustness. RGCN~\cite{rgcn} assigns latent variables to nodes, sampling representations from a Gaussian distribution to enhance robustness and diversity in feature aggregation. Similarly, GNNGuard~\cite{gnnguard} modifies message passing, reducing the influence of suspicious nodes during aggregation.

Although these methods improve GNN robustness, they have disadvantages in practice. Edge-pruning techniques like GCN-Jaccard can oversimplify graphs, discarding important connections. Methods such as Pro-GNN and TSVD rely on assumptions about adversarial perturbations that may not generalize to all attacks. Additionally, many defenses increase computational complexity, making training and inference less scalable for large graphs.

\section{$\mathbf{\beta}$-GNN: Learned-Weighted Ensemble of GNNs and MLP}

Unlike the defense methods mentioned above, we propose ensembling GNNs with a simple MLP, and assigning a learnable parameter, $\beta$, that weighs the output layer embeddings of the GNN. This approach not only enhances the node classification accuracy under perturbation but also provides a tractable parameter to observe the severity of the perturbation during training.

\subsection{Problem Formulation}

A graph is composed of a set of nodes $\mathcal{V}$, edges $\mathcal{E}$, and attributes representing features for entities in the graph, such as node attributes $\boldsymbol{x_i} \in \mathbb{R}^d$. Adversarial attacks on graphs imply perturbing the graph by exploiting the vulnerabilities of GNNs, where an adversarial graph $\mathcal{G'}=(\mathcal{V}, \mathcal{E'}, \boldsymbol X')$ is constructed by implementing small portions of perturbations to the node features 
$\boldsymbol{X} = \{ \boldsymbol{x_i} \}_{i=0}^{n}$
or edge set in the clean graph $\mathcal{G}=(\mathcal{V}, \mathcal{E}, \boldsymbol X)$. Mathematically, adversarial attacks can be framed as an optimization problem where the adversary aims to maximize the loss function $\mathcal{L}(f_\Theta( \mathcal{G'}),  y
)$, where $\Theta$ denotes the GNN parameters and ${y}$ is the prediction label.
The adversary's goal is to find a perturbed graph $\mathcal{G'}$, as defined in \eqref{eq:1}, while adhering to constraints on the perturbation magnitude, such as limitations on feature modifications or the number of altered edges:
\begin{align}
\label{eq:1}
\underset{\mathcal G'=(\mathcal V, \mathcal E', \boldsymbol X')}{\textit{argmax}}& \ \mathcal{L}(\Theta, \mathcal{G'},  y) \\
\text{ s.t. } &|\Delta\mathcal E| \le b_{\mathcal E}, ||\boldsymbol X' - \boldsymbol X|| \le b_{\boldsymbol{X}},
\end{align}
where $\Delta\mathcal E = |\mathcal E| + |\mathcal E'| - 2|\mathcal E\cap \mathcal E'|$ is the number of edges removed or added, $b_{\mathcal E}$ and $b_{\boldsymbol{X}}$ are the perturbation budgets.

In the context of adversarial attacks, the robust optimization framework modifies the traditional objective
\begin{align}
    & \underset{\Theta}{\textit{min}} \ \mathbb{E}_{(\mathcal{G},y) \sim \mathcal{D}} \left[\mathcal L (f_\Theta(\mathcal G), y)\right], \label{eq:2}
\end{align}
to the objective
\begin{align}
    & \underset{\Theta}{\textit{min}} \ \mathbb{E}_{(\mathcal{G},y) \sim \mathcal{D}} \left[\underset{\mathcal G' \in \mathcal P(\mathcal G)}{\textit{max}} \mathcal L (f_\Theta(\mathcal G'), y) \right]. \label{eq:3}
\end{align}
This follows from the definition where \(\mathcal{L} (\Theta, \mathcal{G}, y)\) denotes the loss function, \(\mathcal{D}\) represents the data distribution from which samples \((\mathcal{G}, y)\) are drawn, and \(\mathcal{P}(\mathcal{G})\) is the set of allowable perturbed graphs within a given budget. The term $\max_{\mathcal{G}' \in \Delta(\mathcal{G})} \mathcal{L} (f_\Theta(\mathcal{G}'), y)$ captures the worst-case loss under potential adversarial perturbations. Equation \eqref{eq:1} can thus be interpreted as a graph-specific instance of this formulation.

Robust optimization aims to find model parameters $\Theta$ that minimize the worst-case loss to avoid performance degradation under adversarial conditions.

\subsection{Learned-Weighted Ensembling}

To overcome the poisoning attacks and assess attack severity, we propose ensembling any target GNN with an MLP by calculating the weighted averages of their outputs in the ensemble model, where the weight $\beta$ is learned during the training process. This intuition comes from the proven effectiveness of ensembling~\cite{ensemble} to avoid perturbations and elaborates on how to merge the ensembled models. Considering 
$\Psi$ represents the parameters of MLP, where we have $\hat y_{\text{GNN}} = f_{\Theta}(\mathcal{G})$ and $\hat y_{\text{MLP}}=g_{\Psi}(\boldsymbol{X})$
, the final output of the $\beta$-GNN can be expressed as
\begin{equation}
\label{eq:4}
\hat{y} = \beta \cdot\hat y_{\text{GNN}} + (1-\beta) \cdot \hat y_{\text{MLP}}.
\end{equation}

Following \eqref{eq:3} and \eqref{eq:4}, the $\beta$-GNN's robust optimization problem can be expressed as
\begin{equation}
\label{eq:5}
\begin{aligned}
    \underset{\Theta, \Psi, \beta}{\textit{min}} \ \mathbb{E}_{(\mathcal{G},y) \sim \mathcal{D}} \bigg[ \underset{\mathcal G' \in \mathcal P(\mathcal G)}{\textit{max}} \ 
    \mathcal L& \Big( \beta \cdot f_{\Theta} (\mathcal G') \\
    & + (1 - \beta) \cdot g_{\Psi}(\boldsymbol{X}'), 
    y \Big) \bigg].
\end{aligned}
\end{equation}

First, we consider a single sample and derive the optimal value for $\beta$. By rewriting the loss function $\mathcal{L}$ in \eqref{eq:5}, and deriving it with respect to $\beta$, we express the loss $\mathcal{L}(\beta)$ as a function of $\beta$, as shown in \eqref{eq:7}. This implies that if the predictions of the GNN and the MLP are identical, the derivative of the loss with respect to $\beta$ is always zero. Consequently, $\beta$ does not influence the loss and can take any value. This also indicates that $\beta$ does not affect the predicted value $\hat{y}$.

\begin{align}
    \label{eq:7}
    \frac{\partial \mathcal L}{\partial \beta} &= \frac{\partial \mathcal L}{\partial \hat y}\frac{\partial \hat y}{\partial \beta} = \frac{\partial \mathcal L}{\partial \hat y} \left(f_{\Theta} (\mathcal G')  - g_{\Psi}(\boldsymbol{X}')\right),
\end{align}

In the case where the prediction of GNN and MLP are different, $\frac{\partial \mathcal L}{\partial \beta}=0$ only when $\frac{\partial \mathcal L}{\partial \hat y}=0$, which for many loss functions means $\hat y = y$. Thus, it can be deduced that $\beta$ is only being learned actively when these models generate different predictions.

As adversarial attacks make the GNN's output less reliable by altering the graph structure, the model adjusts $\beta$ to downweight the underlying GNN block and rely more on the output embeddings of the MLP, which primarily works on node features. Similarly, if the node features are under attack, the model expects the opposite behavior. In both cases, comparing the learned $\beta$ values for clean and perturbed cases during training tells practitioners about how likely the data is perturbed or the severity of the perturbation.

\section{Experiments and Results}

\subsection{Experimental Setup}

We conducted the experiments using gradient clipping in PyTorch on a single RTX 5000 GPU with 16 GB of memory. The reported test results correspond to accuracy on the test set when the highest validation accuracy was recorded. For simplicity, we considered 10\% and 20\% perturbation rates for Metattack and 1–5 edge perturbations per node for Nettack.

\paragraph{Datasets}
To evaluate our approach, we run experiments on widely-used benchmark datasets with varying characteristics. Table \ref{tab:data_statistics} presents the key statistics of these datasets, as also reported in~\cite{garnet}. Among these datasets, Cora~\cite{cora} and Pubmed~\cite{pubmed} are citation networks where nodes represent academic papers, edges denote citations, and node features are derived from paper content. Chameleon and Squirrel are web networks extracted from Wikipedia, where nodes represent web pages and edges indicate mutual links between pages~\cite{squirrel_chameleon}. Node features are based on several informative statistics, including average monthly traffic, and text length. These heterophilic datasets are particularly challenging~\cite{hetero_robust} due to their nature, where connected nodes often belong to different classes, such as network traffic graphs in security systems~\cite{hetero_network_sys}, and hardware fault detection graphs~\cite{fault_detection}.

\begin{table}[t]
\centering
\caption{Details of graph datasets used in our experiments.}

\label{tab:data_statistics}
\scriptsize %
\setlength{\tabcolsep}{1pt} %
\renewcommand{\arraystretch}{1.2} %
\resizebox{\linewidth}{!}{
\begin{tabular}{|c|c|c|c|c|c|c|}
\hline
\textbf{Dataset} & \textbf{Type} & \textbf{Homophily Score} & \textbf{\# Nodes} & \textbf{\# Edges} & \textbf{Classes} & \textbf{Features} \\
\hline
Cora & Homophily & 0.80 & 2,485 & 5,069 & 7 & 1,433 \\

Pubmed & Homophily & 0.80 & 19,717 & 44,324 & 3 & 500 \\

Chameleon & Heterophily & 0.23 & 2,277 & 62,792 & 5 & 2,325 \\

Squirrel & Heterophily & 0.22 & 5,201 & 396,846 & 5 & 2,089 \\
\hline
\end{tabular}
}
\vspace{-1mm}
\end{table}

\paragraph{Baselines}
We compare our ensemble approach against three state-of-the-art GNN architectures. The Graph Convolutional Network (GCN)~\cite{gcn} serves as our primary baseline, leveraging first-order spectral graph convolutions through neighborhood aggregation. Second, the Graph Attention Network (GAT)~\cite{gat} enhances GCN by using attention mechanisms to assign different weights to neighbors, improving message passing. Lastly, the Graph PageRank Neural Network (GPRGNN)~\cite{gprgnn} integrates PageRank into GNNs for greater robustness against heterophily and over-smoothing. To ensure a fair comparison, we tune each model’s hyperparameters on the validation set of clean graphs, preventing any implicit adaptation to attacks or noise.

Lastly, we compare $\beta$-GNN against Pro-GNN, GCN-SVD~\cite{gcnsvd}, RGCN, and GCN-Jaccard, to evaluate the effectiveness of the proposed method against other robust GNN models. For heterophilic graphs, GCN-Jaccard implementation returns division by zero error, due to the isolated nodes or regions in the graph. Thus, for Chameleon and Squirrel datasets, we modify GCN-Jaccard so that it assigns zero value to the similarity when similarity cannot be calculated due to the division by zero error. We follow the suggested hyperparameters given in~\cite{deeprobust} for all benchmark models.

\paragraph{Graph adversarial attacks}
To assess the node classification performance of $\beta$-GNN under perturbations, we consider the following attack strategies:

\begin{figure}[htbp]
    \centering
    \includegraphics[width=\columnwidth]{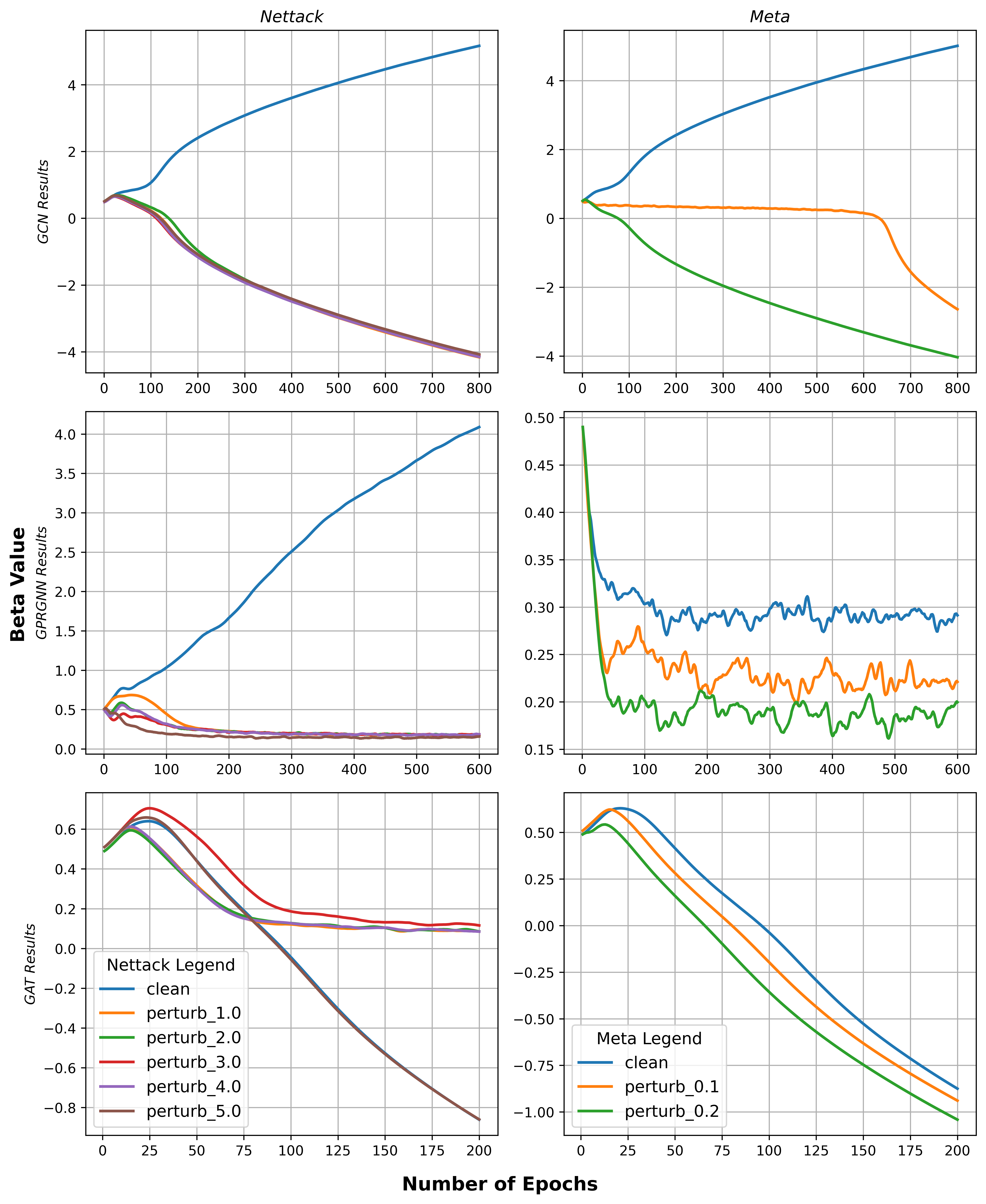}
    \caption{Trajectory of beta values on Pubmed. Left: Nettack. Right: Metattack. Rows: GCN, GPRGNN, GAT (top to bottom).}
    \label{fig:beta_pubmed}
    \vspace*{-4mm}
\end{figure}

\begin{itemize}
    \item \textbf{Untargeted attacks.} We employ a foundational targeted attack method, Mettack, to reduce the performance of GNNs on the entire graph~\cite{meta}.
    \item \textbf{Targeted attacks.} Targeted attacks perform perturbations on specific nodes to mislead GNNs. We followed Nettack~\cite{nettack} as the untargeted attack method and obtained the perturbed graphs from~\cite{prognn, garnet}. For targeted attacks, the test set only includes the target nodes.
\end{itemize}

Adversarial attacks can be categorized into two distinct settings: poisoning, where the graph is perturbed before the GNN is trained, and evasion, where perturbations are applied after the GNN has been trained. Defending against attacks in the poisoning setting is generally more challenging because the altered graph structure directly influences the training process of the GNN~\cite{hetero_robust}. Consequently, our focus is directed toward strengthening model robustness against adversarial attacks in the poisoning setting.

We utilize 10\% of the nodes for training, another 10\% for validation, and the remaining 80\% for the test set as a standard split ratio in robustness studies in GNNs~\cite{prognn}. All perturbations are applied exclusively to the edges, while the node features and labels remain unaffected.

\begin{table*}[t]
\centering
\caption{Node classification accuracy (\%) and standard deviation under Nettack with perturbation budget of 1.0 to 5.0.}
\label{tab:nettack}
\footnotesize

\setlength{\tabcolsep}{2pt} %
\renewcommand{\arraystretch}{1.0} %
\resizebox{0.86\linewidth}{!}{
\begin{tabular}{|c|c|c c|c c|c c|}
\hline
\textbf{Model} & \multicolumn{1}{c|}{\textbf{MLP}} & \multicolumn{2}{c|}{\textbf{GCN}} & \multicolumn{2}{c|}{\textbf{GAT}} & \multicolumn{2}{c|}{\textbf{GPRGNN}} \\ 
\cline{3-8} 
                & \textbf{Vanilla} & \textbf{Vanilla} & \textbf{$\boldsymbol{\beta}$-GNN} & \textbf{Vanilla} & \textbf{$\boldsymbol{\beta}$-GNN} & \textbf{Vanilla} & \textbf{$\boldsymbol{\beta}$-GNN} \\ \hline
Cora-0          & 59.28 ± 2.03 & 80.36 ± 1.71 & \textbf{81.93 ± 1.70}  & 80.00 ± 1.90 & \textbf{81.45 ± 1.52} & \textbf{82.41 ± 214}  & 81.20 ± 2.55 \\ 
Cora-1          & 60.72 ± 4.72 & 75.42 ± 1.90 & \textbf{79.28 ± 1.87}  & \textbf{78.31 ± 1.50} & 77.35 ± 2.71   & 78.80 ± 262 & \textbf{80.24 ± 2.79} \\ 
Cora-2          & 59.88 ± 2.61 & 69.64 ± 1.37 & \textbf{74.46 ± 1.95}  & \textbf{72.29 ± 2.60} & 70.24 ± 3.11  & 74.22 ± 2.21 & \textbf{75.90 ± 1.97} \\ 
Cora-3          & 59.76 ± 4.06 & 64.70 ± 1.80 & \textbf{70.36 ± 0.84}  & 66.14 ± 2.87 & \textbf{67.95 ± 3.07}  & 71.20 ± 2.98 & \textbf{72.41 ± 2.75} \\ 
Cora-4          & 58.67 ± 3.11 & 60.12 ± 1.84 & \textbf{63.49 ± 0.81}  & 60.24 ± 4.13 & \textbf{61.08 ± 4.10}  &  66.39 ± 2.00 & \textbf{69.76 ± 3.43} \\
Cora-5          & 60.24 ± 3.77 & 52.77 ± 1.68 & \textbf{63.98 ± 2.08}  & 55.06 ± 3.73 & \textbf{59.04 ± 4.25 }  & 60.48 ± 3.63 & \textbf{65.06 ± 3.06} \\ 
\hline
Pubmed-0       & 85.91 ± 0.42  & 90.22 ± 0.34 & \textbf{92.53 ± 0.93}  & \textbf{89.95 ± 0.72} & 89.30 ± 0.59 & 91.40 ± 0.91 & \textbf{91.88 ± 1.03} \\ 
Pubmed-1       & 85.59 ± 0.34  & 86.99 ± 0.87 & \textbf{90.65 ± 0.96}  & 87.80 ± 0.72 & \textbf{88.01 ± 1.13} & 88.60 ± 0.83 & \textbf{89.62 ± 0.67} \\
Pubmed-2       & 85.65 ± 0.36  & 85.38 ± 0.49 & \textbf{88.76 ± 1.06}  & 85.00 ± 0.78 & \textbf{86.45 ± 1.16} & 86.40 ± 0.62 & \textbf{87.65 ± 0.84} \\
Pubmed-3       & 85.54 ± 0.40  & 83.12 ± 0.58 & \textbf{86.77 ± 0.68}  & 82.10 ± 1.60 & \textbf{86.88 ± 0.58} & 83.87 ± 0.95 & \textbf{85.81 ± 0.85} \\
Pubmed-4       & 85.91 ± 0.34  & 76.45 ± 0.91 & \textbf{85.00 ± 0.86}  & 79.35 ± 1.05 & \textbf{85.27 ± 0.77} & 80.81 ± 1.13 & \textbf{83.06 ± 1.25} \\
Pubmed-5       & 85.75 ± 0.28  & 68.87 ± 1.23 & \textbf{83.12 ± 1.37}  & 71.67 ± 1.43 & \textbf{84.30 ± 2.64} & 77.31 ± 0.71 & \textbf{83.87 ± 1.52} \\
\hline
Chameleon-0    & 46.46 ± 2.03 & \textbf{78.17 ± 1.07} & 76.83 ± 1.99 & 74.39 ± 1.91 & \textbf{74.88 ± 2.45} & \textbf{75.73 ± 3.07} & 75.61 ± 2.63 \\
Chameleon-1    & 46.59 ± 2.06 & \textbf{73.17 ± 1.29} & 72.44 ± 2.24 & 72.93 ± 2.92 & \textbf{73.54 ± 2.51} & 71.10 ± 1.63 & \textbf{74.02 ± 1.53} \\
Chameleon-2    & 46.46 ± 1.86 & 67.32 ± 1.12 & \textbf{67.93 ± 1.91} & \textbf{70.98 ± 3.14} & 68.66 ± 1.63 & 67.44 ± 2.51 & \textbf{70.37 ± 2.76} \\
Chameleon-3    & 47.20 ± 1.73 & 63.78 ± 0.82 & \textbf{64.27 ± 1.29} & \textbf{68.29 ± 4.88} & 67.07 ± 3.25 & 66.34 ± 2.94 & \textbf{70.61 ± 3.07} \\
Chameleon-4    & 47.32 ± 2.62 & \textbf{63.05 ± 0.59} & 61.95 ± 2.06 & \textbf{65.49 ± 4.19} & 64.51 ± 3.61 & 65.37 ± 2.58 & \textbf{70.00 ± 2.38} \\
Chameleon-5    & 47.80 ± 2.62 & \textbf{61.46 ± 1.03} & 61.10 ± 1.67 & \textbf{64.51 ± 4.12} & 62.56 ± 3.77 & 63.54 ± 5.70 & \textbf{65.37 ± 2.58} \\
\hline     
Squirrel-0     & 28.82 ± 1.82 & \textbf{30.64 ± 2.10} & 29.45 ± 1.43 & \textbf{30.82 ± 6.16} & 28.82 ± 1.22 & \textbf{45.09 ± 1.61}  & 42.36 ± 2.51 \\
Squirrel-1     & 29.82 ± 2.38 & 27.18 ± 1.63 & \textbf{29.73 ± 3.32} & 25.73 ± 3.56 & \textbf{29.82 ± 1.12} & \textbf{43.55 ± 1.89} & 40.82 ± 1.94 \\
Squirrel-2     & 31.18 ± 2.19 & 29.36 ± 1.14 & \textbf{29.39 ± 2.87} & 24.45 ± 3.47 & \textbf{29.27 ± 1.47} & \textbf{42.64 ± 2.55} & 41.18 ± 1.72 \\
Squirrel-3     & 28.18 ± 3.78 & \textbf{27.91 ± 2.88} & 27.36 ± 1.98 & 23.55 ± 6.05 & \textbf{29.64 ± 1.50} & 42.09 ± 2.78 & \textbf{43.00 ± 3.98} \\
Squirrel-4     & 29.55 ± 2.79 & \textbf{29.55 ± 0.64} & 28.09 ± 2.44 & 24.09 ± 3.35 & \textbf{30.36 ± 0.98} & \textbf{42.45 ± 2.54} & 41.45 ± 1.50 \\
Squirrel-5     & 28.27 ± 3.62 & \textbf{29.55 ± 2.88} & 28.82 ± 1.92 & 21.82 ± 3.51 & \textbf{28.64 ± 1.83} & 42.55 ± 1.47 & \textbf{44.45 ± 3.25} \\
\hline
\end{tabular}
}
\end{table*}

\begin{table*}[t]
\centering
\caption{Node classification accuracy (\%) and standard deviation under Metattack with perturbation ratio 0.0 to 0.2.}
\label{tab:meta}
\footnotesize

\setlength{\tabcolsep}{2pt} %
\renewcommand{\arraystretch}{1.0} %
\resizebox{0.86\linewidth}{!}{
\begin{tabular}{|c|c|c c|c c|c c|}
\hline
\textbf{Model} & \multicolumn{1}{c|}{\textbf{MLP}} & \multicolumn{2}{c|}{\textbf{GCN}} & \multicolumn{2}{c|}{\textbf{GAT}} & \multicolumn{2}{c|}{\textbf{GPRGNN}} \\ 
\cline{3-8} 
                & \textbf{Vanilla} & \textbf{Vanilla} & \textbf{$\boldsymbol{\beta}$-GNN} & \textbf{Vanilla} & \textbf{$\boldsymbol{\beta}$-GNN} & \textbf{Vanilla} & \textbf{$\boldsymbol{\beta}$-GNN} \\ \hline
Cora-0      & 62.27 ± 1.77 & \textbf{83.64 ± 0.81} & 83.18 ± 0.85  & \textbf{83.71 ± 0.56} & 83.47 ± 0.72 & \textbf{83.69 ± 0.71} & 83.51 ± 0.73 \\ 
Cora-0.1    & 63.52 ± 1.32 & 74.78 ± 0.94 & \textbf{76.73 ± 0.38}  & \textbf{76.80 ± 0.75} & 76.55 ± 0.99 & 76.87 ± 0.88 & \textbf{79.07 ± 0.79}  \\ 
Cora-0.2    & 63.30 ± 1.74 & 58.73 ± 0.71 & \textbf{68.56 ± 2.36}  & 60.60 ± 2.40 & \textbf{70.26 ± 1.73} & 69.15 ± 3.44 & \textbf{74.75 ± 0.56}  \\ 

\hline
Pubmed-0    & 84.43 ± 0.24 & 87.14 ± 0.05 & \textbf{87.76 ± 0.28} & 85.65 ± 0.22 & \textbf{87.15 ± 0.23} & \textbf{88.43 ± 0.27} & 88.37 ± 0.14 \\ 
Pubmed-0.1  & 84.47 ± 0.22 & 81.20 ± 0.08 & \textbf{86.85 ± 0.14} & 80.46 ± 0.35 & \textbf{86.15 ± 0.12} & \textbf{87.42 ± 0.14} & 87.39 ± 0.18 \\
Pubmed-0.2  & 84.34 ± 0.22 & 77.17 ± 0.19 & \textbf{86.34 ± 0.18} & 76.36 ± 0.24 & \textbf{85.18 ± 0.24} & 86.72 ± 0.25 & \textbf{86.81 ± 0.22} \\
\hline
Chameleon-0     & 48.45 ± 0.75 & \textbf{67.37 ± 0.49} & 64.90 ± 0.98 & \textbf{65.25 ± 0.78} & 61.88 ± 1.30 & \textbf{68.84 ± 0.49} & 68.82 ± 0.80 \\
Chameleon-0.1   & 47.75 ± 0.80 & 53.44 ± 0.92 & \textbf{54.88 ± 1.27} & 52.39 ± 1.70 & \textbf{52.63 ± 1.22} & 62.23 ± 1.03 & \textbf{62.93 ± 0.53} \\
Chameleon-0.2   & 48.34 ± 0.80 & 51.51 ± 0.84 & \textbf{52.71 ± 1.12} & 48.75 ± 1.41 & \textbf{50.61 ± 1.62} & 59.61 ± 1.04 & \textbf{60.28 ± 1.03} \\

\hline     
Squirrel-0     & 33.65 ± 0.89 & 58.40 ± 0.77  & \textbf{59.58 ± 0.71}  & \textbf{46.04 ± 2.25} & 42.67 ± 1.44  & 53.79 ± 0.67 & \textbf{53.94 ± 0.66}  \\
Squirrel-0.1   & 33.47 ± 0.78 & 47.45 ± 0.70  & \textbf{48.64 ± 0.75}  & \textbf{41.84 ± 0.88} & 40.97 ± 2.18  & 46.05 ± 1.07 & \textbf{47.17 ± 0.55}  \\
Squirrel-0.2   & 33.78 ± 0.77 & 42.65 ± 0.81  & \textbf{42.97 ± 2.81}  & 39.29 ± 1.83 & \textbf{39.43 ± 2.28}  & 43.17 ± 0.87 & \textbf{44.05 ± 0.81}  \\
\hline
\end{tabular}
}
\end{table*}

\begin{table*}[t]
\centering
\caption{Benchmark of defense methods against Nettack. \textit{OOM} refers to out-of-memory error due to the high GPU memory requirements of particular models.}
\label{tab:benchmark_nettack}
\small
\resizebox{\linewidth}{!}{
\begin{tabular}{l c c c c c c c c}
\hline
\multirow{2}{*}{\textbf{Model}} & \multicolumn{2}{c}{\textbf{Cora}} & \multicolumn{2}{c}{\textbf{Pubmed}} & \multicolumn{2}{c}{\textbf{Chameleon}} & \multicolumn{2}{c}{\textbf{Squirrel}} \\
\cline{2-9}
& Clean & Perturbed & Clean & Perturbed & Clean & Perturbed & Clean & Perturbed \\
\hline
GCN-Vanilla & 80.36 ± 1.71 & 52.77 ± 1.68  & 90.22 ± 0.34 & 68.87 ± 1.23  & \textbf{78.17 ± 1.07}  & 61.46 ± 1.03 & 30.64 ± 2.10  & 29.55 ± 2.88  \\
GCN-SVD & 81.81 ± 1.05 & 60.72 ± 1.41 & \textit{OOM} & \textit{OOM}  & 65.00 ± 0.59 & 61.95 ± 1.12 & 26.45 ± 2.20  & 22.64 ± 1.79 \\
GCN-Jaccard & 78.43 ± 1.92 & 64.10 ± 3.76  & 90.65 ± 0.38  & 70.43 ± 2.04 & 65.85 ± 1.82 & 60.61 ± 1.29  & 25.91 ± 1.56 & 24.27 ± 1.49 \\
RGCN & 80.36 ± 1.40 & 56.63 ± 2.48  & \textit{OOM} & \textit{OOM} & 67.80 ± 2.52 & 50.85 ± 4.03 & 32.27 ± 2.62  & 16.27 ± 1.25  \\
Pro-GNN & \textbf{84.46 ± 1.44} & 58.55 ± 2.14 & \textit{OOM} & \textit{OOM} & 76.46 ± 2.57 & 61.71 ± 1.54 & 37.55 ± 1.87 & 24.36 ± 6.55  \\
$\beta$-GNN & 81.20 ± 2.55 & \textbf{65.06 ± 3.06}  & \textbf{91.88 ± 1.03}  & \textbf{83.87 ± 1.52}  & 75.61 ± 2.63 & \textbf{65.37 ± 2.58} & \textbf{42.36 ± 2.51}  & \textbf{44.45 ± 3.25} \\
\hline
\end{tabular}
}
\end{table*}

\begin{table*}[t]
\centering
\caption{Benchmark of defense methods against Metattack.}
\vspace{-2mm}
\label{tab:benchmark_meta}
\footnotesize
\resizebox{\linewidth}{!}{
\begin{tabular}{l c c c c c c c c}
\hline
\multirow{2}{*}{\textbf{Model}} & \multicolumn{2}{c}{\textbf{Cora}} & \multicolumn{2}{c}{\textbf{Pubmed}} & \multicolumn{2}{c}{\textbf{Chameleon}} & \multicolumn{2}{c}{\textbf{Squirrel}} \\
\cline{2-9}
& Clean & Perturbed & Clean & Perturbed & Clean & Perturbed & Clean & Perturbed \\
\hline
GCN-Vanilla & 83.64 ± 0.81 & 58.73 ± 0.71  & 87.14 ± 0.05 & 77.17 ± 0.19 & 67.37 ± 0.49 & 51.51 ± 0.84 & \textbf{58.40 ± 0.77} & 42.65 ± 0.81 \\
GCN-SVD & 78.37 ± 3.49 & 61.45 ± 1.64 & \textit{OOM} & \textit{OOM} & 47.81 ± 0.31 & 37.72 ± 1.36 & 31.96 ± 0.48 & 23.01 ± 0.76 \\
GCN-Jaccard & 80.73 ± 0.61 & 74.07 ± 0.87 & 87.08 ± 0.07 & 78.08 ± 0.11 & 54.53 ± 0.46 & 48.03 ± 0.65 & 35.83 ± 0.73 & 34.30 ± 0.57 \\
RGCN & 83.47 ± 0.34 & 57.71 ± 0.43 & \textit{OOM} & \textit{OOM} & 56.64 ± 0.67 & 41.93 ± 1.43 & 35.96 ± 0.93  & 28.79 ± 0.74  \\
Pro-GNN & \textbf{84.72 ± 0.36} & 57.11 ± 0.06 & \textit{OOM} & \textit{OOM} & 66.54 ± 1.64 & 54.88 ± 0.86 & 48.21 ± 4.29 & 30.05 ± 0.71  \\
$\beta$-GNN & 83.69 ± 0.71 & \textbf{74.75 ± 0.56} & \textbf{88.37 ± 0.14} & \textbf{86.81 ± 0.22} & \textbf{68.82 ± 0.80} & \textbf{60.28 ± 1.03} & 53.94 ± 0.66 & \textbf{44.05 ± 0.81}  \\
\hline
\end{tabular}
}
\end{table*}

\subsection{Results}

Table \ref{tab:nettack} reports the averaged accuracy and standard deviation over 10 training runs when graphs are under targeted attack, whereas Table \ref{tab:meta} reports the results under untargeted attack. For homophilic graphs, where similar nodes are linked, it can be deduced that $\beta$-GNN gains up to 14.25\% of test accuracy, compared to the baseline GNN models. However, in the case of heterophilic graphs, where dissimilar nodes are connected, $\beta$-GNN does not lead to a significant improvement in performance.  These results suggest that while $\beta$-GNN offers benefits across graph types, the challenges posed by weak label correlations in heterophilic graphs require further investigation. Nonetheless, it either enhances test accuracy or performs comparably to the baseline models.

Furthermore, the learnable weight of the averaging variable, $\beta$, is tracked during training, as depicted by Fig. \ref{fig:beta_pubmed}. The intuition that $\beta$ can serve as an effective parameter for assessing attack severity is supported by the observation that, in most cases, $\beta$ values are distinguishable for clean graphs. This trajectory also shows that the severity levels can be distinguishable either, considering how $\beta$ varied between different perturbation rates, i.e. for all models under Metattack in Fig. \ref{fig:beta_pubmed}. The early estimation of $\beta$ allows practitioners to identify the presence and severity of adversarial attacks during the initial stages of the training process, enabling proactive measures to improve model robustness before training is completed.

Fig. \ref{fig:beta_pubmed} also shows a result that contradicts the intuition behind learning the $\beta$ values, when GAT is used as the backbone model under Nettack attack on the Pubmed dataset. This is particularly interesting because, despite $\beta$-GNN showing significant performance improvement over vanilla GAT (84.30\% vs 71.67\% accuracy for 5-edge perturbations), the $\beta$ values do not show a clear separation between clean and perturbed cases. This suggests that while the ensemble model effectively enhances model robustness, the mechanism of improvement might be different from other backbone architectures. Rather than clearly down-weighting the GAT component under attack, the model might be finding a more complex integration strategy between GAT and MLP that leverages GAT's attention mechanism in conjunction with MLP's feature processing, even under perturbation.

In addition to the experiments that compare the proposed method against the vanilla backbone models, we compare $\beta$-GNN against other defense methods. Table \ref{tab:benchmark_nettack} and \ref{tab:benchmark_meta} present the benchmark results, where GPRGNN is used as the backbone for the proposed method due to its excellent performance on heterophilic datasets. For simplicity, we only present results for perturbed graphs with a perturbation budget of 5 edges per target node for Nettack, and a perturbation ratio of 20\% for Metattack in Table \ref{tab:benchmark_nettack} and \ref{tab:benchmark_meta}.

\subsection{Complexity Analysis}

Computational efficiency is crucial to real-world applicability of defense methods. Herein, we analyze the time and space complexity of $\beta$-GNN and other defense models. Let $T$ be the number of iterations in Pro-GNN’s optimization process, $N$ be the number of nodes, $M$ be the number of edges, $K$ be the number of graph convolutional layers, $F$ be the feature dimension, $H$ be the hidden layer dimension, and $k$ be the number of top singular vectors retained in GCN-SVD.

\begin{itemize}
    
    \item \textbf{Pro-GNN} introduces significant computational overhead through iterative structure learning, rendering it impractical for large-scale graphs. With our experimental setup, Pro-GNN takes almost 2 days to train on Squirrel graph to produce the reported results. Its time complexity is $O(T(N^3 + KMN)$, and its space complexity is $O(N^2)$.
    \item \textbf{GCN-Jaccard} is comparably better than other benchmarks in terms of its time complexity. However, the quadratic edge similarity computation limits its scalability for dense graphs. Furthermore, its performance on heterophilic graphs is even lower than GCN-Vanilla, according to Table \ref{tab:benchmark_nettack} and \ref{tab:benchmark_meta}. Its time complexity and space complexity can be expressed as $O(M^2 + KMN)$ and $O(M+NF)$, respectively.
    \item \textbf{RGCN} introduces additional computational complexity through Gaussian-based graph learning. Along with Pro-GNN and GCN-SVD, RGCN cannot scale to larger graphs, as evidenced by our evaluations on the Pubmed dataset, where it results in an out-of-memory error. The time complexity of RGCN is $O(KMN + KN^2)$, whereas its space complexity is $O(N^2 + M)$
    \item \textbf{GCN-SVD} introduces cubic complexity due to applying singular value decomposition, severely restricting scalability. Computing the full matrix eigendecomposition requires $O(N^3)$ computation complexity and results in a high time complexity $O(N^3+KMN)$ for GCN-SVD. Its space complexity is $O(kN+M)$, where $k$ needs to increase as the graph size grows.
    \item \textbf{$\beta$-GNN} method achieves linear scalability by avoiding expensive graph structure preprocessing. The time complexity of the proposed method is $O(KMH + FNH)$, while its space complexity is $O(M+NF)$.
\end{itemize}
\makeatletter
\renewcommand{\@nobreaktrue}{}
\makeatother

We observe that cubic complexity models (Pro-GNN, GCN-SVD) become impractical for graphs with $N > 10,000$ nodes. Similarly, quadratic models (RGCN) face significant performance degradation. In contrast, $\beta$-GNN, as a linear complexity model, maintains consistent performance across varying graph scales.

\section{Conclusion}

In this work, we introduced $\beta$-GNN, a novel approach to enhance GNN robustness by dynamically weighting the contribution of a base GNN model and an MLP through a learned $\beta$ parameter. This ensemble method not only improves resilience against adversarial attacks but also provides an interpretable measure of data perturbation severity. Our experiments demonstrate the effectiveness of $\beta$-GNN in improving node classification accuracy under attack, particularly in preserving performance on unperturbed data structures. The linear computational complexity of $\beta$-GNN offers a significant advantage for scalability in large-scale applications.

While $\beta$-GNN demonstrates promising results, it is not without limitations. Specifically, while the learned $\beta$ values often provide a clear distinction between clean and perturbed data instances, this separation is not always guaranteed. In some cases, the tracks of $\beta$ values for clean and perturbed data can intertwine, making it challenging to definitively distinguish between them. Future work will address this limitation by investigating methods to improve the separation of $\beta$ value tracks, potentially through incorporating additional features or constraints during the learning process. This could involve exploring more sophisticated regularization techniques or examining the influence of different base GNN architectures on the behavior of $\beta$.

\begin{acks}
This paper was supported by the Swarmchestrate project of the European Union’s Horizon 2023 Research and Innovation programme under grant agreement no. 101135012.
\end{acks}

\bibliographystyle{ACM-Reference-Format}
\bibliography{main}

\end{document}